\documentclass[10pt,twocolumn,letterpaper]{article}

\usepackage{cvpr}           
\definecolor{cvprblue}{rgb}{0.21,0.49,0.74}

\usepackage{multirow}
\definecolor{emerald}{RGB}{80, 200, 120}
\definecolor{coral}{RGB}{255, 127, 80}
\definecolor{teal}{RGB}{0, 128, 128}
\definecolor{goldenrod}{RGB}{218, 165, 32}

\usepackage[pagebackref,breaklinks,colorlinks,allcolors=cvprblue]{hyperref}

\usepackage[table]{xcolor}
\usepackage{tcolorbox}

\title{Constructing and Interpreting Digital Twin Representations for Visual Reasoning via Reinforcement Learning}

\author{Yiqing Shen, Mathias Unberath\\
Department of Computer Science, Johns Hopkins University\\
{\tt\small \{yshen92, unberath\}@jhu.edu}
}

\begin{document}
\maketitle
\begin{abstract}
Visual reasoning may require models to interpret images/videos and respond to implicit text queries across diverse output formats, from pixel-level segmentation masks to natural language descriptions. 
Existing approaches rely on supervised fine-tuning with task-specific architectures.
For example, reasoning segmentation, grounding, summarization, and visual question answering each demand distinct model designs and training, preventing unified solutions and limiting cross-task/-modality generalization.
Hence, we propose DT-R1, a reinforcement learning framework that trains large language models (LLMs) to construct digital twin (DT) representations of complex multi-modal visual inputs and then reason over these high-level representations as a unified approach to visual reasoning.
Specifically, we train DT-R1 using GRPO with a novel reward that validates both structural integrity and output accuracy.
Evaluations in six visual reasoning benchmarks, covering two modalities and four task types, demonstrate that DT-R1 consistently achieves improvements over state-of-the-art task-specific models. 
DT-R1 opens a new direction where visual reasoning emerges from reinforcement learning on with DT representations.
\end{abstract}    
\section{Introduction}
\label{sec:intro}

Visual reasoning supports embodied AI systems to interpret visual data (\textit{e}.\textit{g}., images or videos) and respond to implicit text queries in various applications, such as robotics and autonomous driving~\cite{gao2023assistgpt}.
Previous work on visual reasoning has primarily focused on reasoning segmentation~\cite{lisa,visa}, where targeted objects are segmented based on implicit text queries like ``\textit{segment the object used to hold hot beverages}'' rather than explicit expressions~\cite{referringsegmentation} such as ``\textit{segment the coffee cup}'' or point prompts~\cite{sam1,sam2}.
However, real-world applications may require diverse output formats beyond pixel-level segmentation masks~\cite{miccai,survey}.
For instance, assistive systems for visually impaired users need textual descriptions of scenes rather than masks or bounding box; while it is more efficient to use bounding box in grounding.
Reasoning visual tasks (RVTs)~\cite{rvtbench} then expand on reasoning segmentation to a diverse family of visual reasoning problems, including reasoning grounding~\cite{zhu2024scanreason} for bounding box generation, reasoning summarization for natural language descriptions, reasoning visual question answering (VQA)~\cite{zakari2022vqa} for textual responses, \textit{etc}.

Yet, current approaches for RVTs face three limitations. 
First, RVT methods such as LISA~\cite{lisa}, VISA~\cite{visa}, and ScanReason~\cite{zhu2024scanreason} encode visual features as tokens for multimodal LLM processing. 
This token-based encoding breaks apart continuous spatial-temporal relationships into discrete tokens, causing the loss of fine-grained geometric information necessary for reasoning, resulting in suboptimal performance~\cite{jit,position}.
Second, most existing RVT methods rely on supervised fine-tuning (SFT), which produces output without explicit reasoning chains~\cite{cot}. 
This prevents decision verification from the users' side, and limits their ability to generalize beyond training distributions as demonstrated by poor performance on out-of-domain data~\cite{segzero,miccai}.
Finally, existing RVT methods remain specific to task types or data modalities, where each RVT requires its own specialized model architecture and training procedure.
These limitations motivate the need for a unified framework that preserves spatial-temporal information while learning reasoning capabilities without task-specific architectural modifications.

Recent work on just-in-time digital twins (DT)~\cite{jit} shows that replacing token embedding with DT representations can preserve spatial-temporal continuity for reasoning segmentation in both image and video~\cite{miccai}, but it is performed in a zero-shot manner.
Consequently, to bridge the gap, we propose DT-R1, a reinforcement learning (RL) framework that trains LLMs to construct and reason over task-specific DT representations, allowing diverse RVTs across image and video to be handled within a single unified model through structured rollout sequences. 
Specifically, DT-R1 trains LLMs to analyze implicit text queries and autonomously generate plans that specify which vision foundation models to apply for DT representation construction, thereby eliminating the need for manual effort in domain-specific design processes. 
Subsequently, the LLM learns to perform iterative reasoning over these DT representations by generating explicit CoTs~\cite{cot} and executable Python code for qualitative reasoning when required, before identifying the appropriate task type and producing the corresponding output format.
Rather than relying on SFT with annotated reasoning chain, DT-R1 gains reasoning capabilities in diverse domains without requiring task-specific tuning through pure RL through Group-Relative Policy Optimization (GRPO)~\cite{grpo}, together with a novel reward that combines format validation for structured outputs and accuracy validation for the final output.

The major contributions are three-fold.
First, we propose DT-R1, an RL framework that trains a single LLM to handle diverse RVTs including reasoning segmentation, grounding, summarization, and visual question answering across two visual modalities, as a unified solution to eliminate the need for task-specific architecture design and tuning. 
Specifically, it designs a structured rollout that trains the LLM to generate DT representation construction plans specifying vision foundation model dependencies, explicit reasoning chains, executable Python code for spatial-temporal computations, task-type identification, and task-appropriate final responses. 
Second, we introduce a novel rule-based reward for DT-R1 training that takes into account the output format and accuracy, including the validity of the DT representation construction plan and the executability of the code.
Third, we demonstrate that pure RL enables the development of reasoning capabilities without requiring explicit supervision on visual data with the self-constructed DT representations, which suggests a new potential venue for scaling.

\section{Related Works}

\paragraph{Reasoning Visual Tasks (RVTs)}
Reasoning segmentation aims to identify the target object and generates the corresponding pixel-level masks by interpreting implicit text queries through multi-step reasoning~\cite{survey}.
Reasoning segmentation methods can be classified into two primary paradigms.
End-to-end methods, such as LISA~\cite{lisa} and VISA~\cite{visa}, employ specialized tokens such as \texttt{<seg>} to integrate the perception and reasoning capabilities of multimodal LLMs (MLLMs) with the segmentation abilities of separate decoders, but may encounter catastrophic forgetting during fine-tuning.
Disentangled architectures, such as LLM-Seg~\cite{llmseg}, CoReS~\cite{cores}, LLaVASeg~\cite{llavaseg} and Seg-Zero~\cite{segzero}, address these limitations by maintaining the separation between reasoning and segmentation by transforming the reasoning output into explicit prompts for segmentation networks~\cite{sam1,sam2} at the cost of additional computational complexity.
Just-in-time DT representation has further disentangled perception from reasoning~\cite{jit} by constructing structured DT representation that preserve fine-grained semantic, spatial, and temporal information, while enabling LLMs to generate executable code for complex reasoning operations in a zero-shot manner, eliminating the need for LLM fine-tuning.
Since reasoning segmentation remains limited to mask generation, reasoning visual tasks (RVTs)~\cite{rvtbench} are introduced as a unified formulation.
Specifically, it extends the reasoning paradigm beyond segmentation to encompass multiple output formats, including bounding box generation, natural language summarization, and visual question answering, while preserving the concept of processing implicit text queries.

\paragraph{Digital Twin Representation}
Digital twins (DTs) in the context of foundation models have been defined as ``\textit{outcome-driven virtual replicas of physical processes that capture task-specific entities and their interactions to analyze and optimize their real-world counterparts}''~\cite{position}. 
DT representations are the building blocks of these virtual DT replicas~\cite{position}, usually comprising structured data formats that encode semantic, spatial, and temporal relationships extracted from raw input. 
In terms of RVTs, previous work has shown the success of DT representation as an intermediate layer between visual perception and reasoning~\cite{jit}, where LLMs dynamically plan the construction of such representations from video using specialized vision foundation models like SAM2~\cite{sam2}, DepthAnything2~\cite{depthanything}, OWLv2~\cite{owlv2} and DINOv2~\cite{dinov2}, based on query.
%
%

\paragraph{Reinforcement Learning for LLMs}
RL is a training paradigm for LLMs to gain reasoning capabilities, as shown by DeepSeek-R1~\cite{r1} with GRPO. 
Similarly, ToRL~\cite{torl} and Search-R1~\cite{searchr1} have shown that RL training can further improve mathematical reasoning or LLM search-based reasoning by iterative reward optimization without requiring explicit supervision.
Seg-Zero~\cite{segzero} and Seg-R1~\cite{segr1} applied pure RL to visual reasoning, achieving superior out-of-domain generalization in terms of reasoning segmentation.
ViGoRL achieves spatially grounded visual reasoning via RL by anchoring each reasoning step to specific visual coordinates~\cite{vigorl}.
However, existing RL-based approaches remain constrained to text-only reasoning domains or individual reasoning visual tasks, whereas our DT-R1 establishes an integration of RL training with DT representations to enable unified reasoning across diverse RVTs.
\section{Methods}

\begin{figure*}[t!]
\centering
\includegraphics[width=\linewidth]{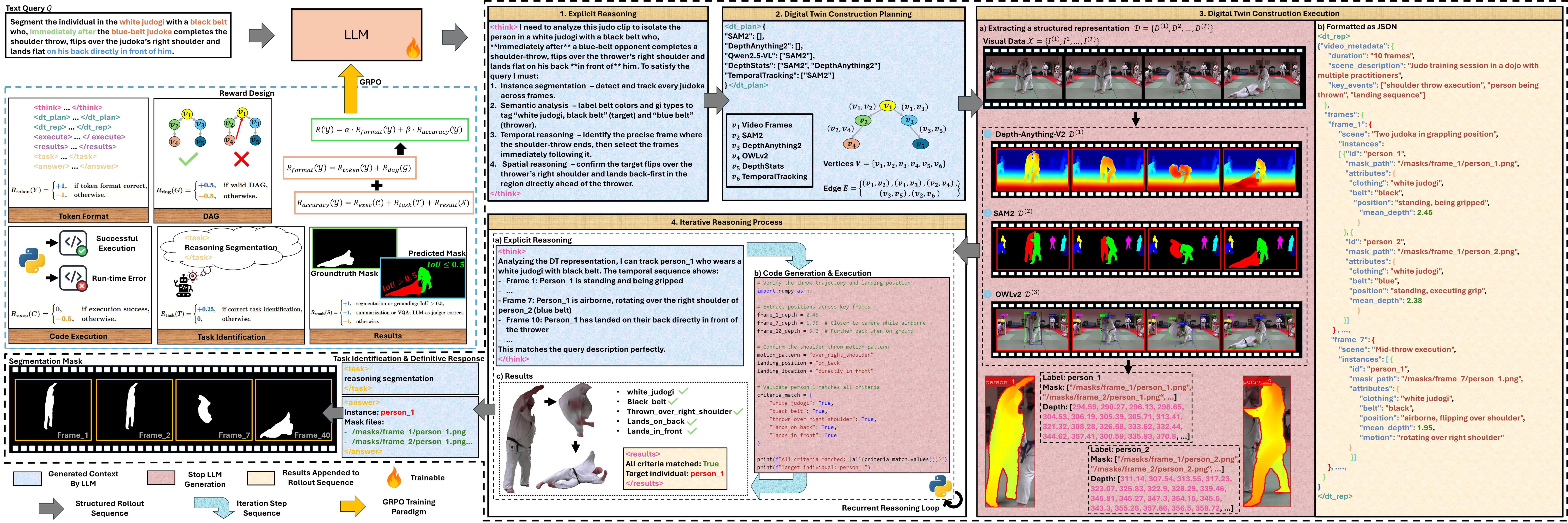} 
\caption{
Overall framework of DT-R1 for unified reasoning visual tasks. Given an implicit text query and visual input (image or video), DT-R1 generates a structured rollout sequences through reinforcement learning. 
}
\label{fig:method}
\end{figure*}

\paragraph{Oveview}
We propose DT-R1, an RL framework that trains LLMs to construct and reason over DT representations as a unified method for reasoning visual tasks \cite{rvtbench} in the context of both image and video modalities, as shown in Fig.~\ref{fig:method}. 
Rather than relying on token representation of visual features~\cite{llava} or supervised fine-tuning with explicit reasoning data, DT-R1 learns to bridge the gap between high-level text query reasoning and low-level visual perception through DT representations \cite{jit,miccai}.
%
%

\paragraph{Structured Rollout Sequence}
DT-R1 trains LLM to plan, construct, and reason on DT representations from the given implicit text query $Q$ via structured output. 
Specifically, it begins the generation of the output sequence by explicit reasoning within \textcolor{magenta}{\texttt{<think>}} and \textcolor{magenta}{\texttt{</think>}} tokens, where the LLM analyzes the requirements of the given text query $Q$ and identifies the types of information needed. 
Afterwards, the LLM then generates the DT representation construction plan within specialized tokens a to specify which vision foundation models should be applied and establish their execution dependencies in terms of a directed acyclic graph (DAG) $\mathcal{G}$. 
The DT representation construction plan is then executed by applying the specified vision foundation models to the given visual data $\mathcal{X} = \{I^{(1)}, I^{(2)}, \ldots, I^{(T)}\}$, where $T$ represents the number of frames and $T=1$ for the image modality.
The planned DT construction is then executed, applying the specified vision foundation models to extract relevant visual information from the input visual data. 
The resulting structured representation $\mathcal{D} = \{D^{(1)}, D^{(2)}, \ldots, D^{(T)}\}$ is formatted as JSON and integrated into the rollout sequence within \textcolor{teal}{\texttt{<dt\_rep>}} and \textcolor{teal}{\texttt{</dt\_rep>}} tokens. 
Upon completion of the DT construction, DT-R1 trains LLM to enter an iterative reasoning process to address the reasoning visual task.
This begins with another round of explicit reasoning within \textcolor{magenta}{\texttt{<think>}} and \textcolor{magenta}{\texttt{</think>}} tokens, where the LLM reasons over the self-constructed DT representation.
When computational operations are required in spatial or temporal reasoning, LLM generates executable Python code within \textcolor{violet}{\texttt{<execute>}} and \textcolor{violet}{\texttt{</execute>}} tokens, using libraries such as OpenCV. 
The execution results are also appended to the rollout sequence within the \textcolor{violet}{\texttt{<results>}} and \textcolor{violet}{\texttt{</results>}} tokens.
This iterative reasoning process continues until the termination conditions are met: either the maximum token limit is reached, or the model generates a final response by determining the appropriate task type and output format within \textcolor{goldenrod}{\texttt{<task>}} and \textcolor{goldenrod}{\texttt{</task>}} tokens, specifying whether the response should contain segmentation masks, bounding boxes, natural language summaries, or textual answers. 
Finally, the definitive response is then generated within \textcolor{goldenrod}{\texttt{<answer>}} and \textcolor{goldenrod}{\texttt{</answer>}} tokens.
The complete structured output sequence can be formulated as:
\begin{equation}
\footnotesize
\nonumber
\begin{aligned}
\mathcal{Y} = \langle &
\textcolor{magenta}{\texttt{<think>}}\mathcal{R}_0\textcolor{magenta}{\texttt{</think>}}, \textcolor{teal}{\texttt{<dt\_plan>}}\mathcal{G}\textcolor{teal}{\texttt{</dt\_plan>}}, \\
&\textcolor{teal}{\texttt{<dt\_rep>}}\mathcal{D}\textcolor{teal}{\texttt{</dt\_rep>}}, (\textcolor{magenta}{\texttt{<think>}}\mathcal{R}_i\textcolor{magenta}{\texttt{</think>}},\\
& \textcolor{violet}{\texttt{<execute>}}\mathcal{C}_i\textcolor{violet}{\texttt{</execute>}},\phantom{(}\textcolor{violet}{\texttt{<results>}}\mathcal{O}_i
\textcolor{violet}{\texttt{</results>}})_{i=1}^{m}, \\
&
\textcolor{goldenrod}{\texttt{<task>}}\mathcal{T}\textcolor{goldenrod}{\texttt{</task>}}, \textcolor{goldenrod}{\texttt{<answer>}}\mathcal{S}\textcolor{goldenrod}{\texttt{</answer>}} \rangle
\end{aligned}
\end{equation}
where $\mathcal{R}_0$ represents the initial analysis, $\mathcal{G}$ denotes the construction plan in terms of DAG, $\mathcal{D}$ is the DT representation, $\mathcal{R}_i$, $\mathcal{C}_i$, and $\mathcal{O}_i$ represent the results of reasoning, code, and execution in iteration $i$, $\mathcal{T}$ specifies the type of task, $\mathcal{S}$ contains the final answer, and $m$ indicates the number of iterative reasoning cycles performed.

\paragraph{Digital Twin Representation Planning and Construction}
This plan for DT representation construction is formulated by a text-encoded DAG~\cite{shi2025reinforcement} $\mathcal{G} = (V, E)$ in tokens \textcolor{teal}{\texttt{<dt\_plan>}} and \textcolor{teal}{\texttt{</dt\_plan>}}, which defines the execution sequence and dependencies for the required vision foundation models.
The graph structure comprises vertices $V = \{v_1, v_2, \ldots, v_K\}$ representing the selected vision foundation models and edges $E = \{(v_i, v_j) | v_i, v_j \in V\}$ denoting directed dependencies between models. 
Each edge $(v_i, v_j) \in E$ establishes that the computational output of model $v_i$ serves as a pre-requisite for model $v_j$. 
During both training and inference, the LLM acquires knowledge of available vision foundation models through prompts that detail each model's specific capabilities, input format requirements, and output specifications.

For example, SAM2~\cite{sam1,sam2} typically functions as a primary computational node for instance segmentation, generating mask collections $\mathcal{M}^{(t)} = \{m_i^{(t)}\}_{i=1}^{N^{(t)}}$ where each $m_i^{(t)}$ represents a binary segmentation mask for the instance of the object $i$ in the temporal frame $t$, with $N^{(t)}$ indicating the total number of detected instances.
To enable processing by LLM, only the path of the mask collections is encoded in the DT representation instead of the mask itself.
DepthAnything2~\cite{depthanything} operates as a dependent computational node that processes video frame input to generate dense depth maps $\mathcal{Z}^{(t)}$, while subsequent depth statistics computation operations depend on both the SAM2-generated masks and DepthAnything2 depth output for spatial analysis.
Following previous work \cite{rvtbench,jit}, VLMs such as Qwen2.5-VL~\cite{qwen} function as semantic analysis nodes that maintain partial dependencies on SAM2 outputs to facilitate multi-level semantic information extraction across instance, frame, and video levels.
In this case, the DAG is represented in text format as a structured dependency specification: \textit{\{"SAM2": [], "DepthAnything2": [], "DepthStats": ["SAM2", "DepthAnything2"], "SemanticAnalysis": ["SAM2"]\}}, where each key is a vision foundation model (\textit{i}.\textit{e}., vertex in the DAG) and the corresponding list specifies its prerequisite dependencies (\textit{i}.\textit{e}., the edges).

When \textcolor{teal}{\texttt{</dt\_plan>}} is detected in the rollout sequence, the text
generation is paused for DT representation generation.
Then, the resulting DT representation $\mathcal{D}$ is inserted between the \textcolor{teal}{\texttt{<dt\_rep>}} and \textcolor{teal}{\texttt{</dt\_rep>}} tokens as a JSON structure that encompasses three levels. 
The video-level metadata captures global scene descriptions and temporal information throughout the video sequence. 
Frame-level information provides scene and spatial descriptions for each timestamp $t$. 
Instance-level attributes contain detailed masks, depth statistics, semantic descriptions, and visual features for individual objects. 
For example, for each instance $i$, depth statistics $d_i^{(t)} = \{\mathcal{Z}^{(t)}(p) | p \in m_i^{(t)}\}$ are computed on all pixels within the instance mask, incorporating mean depth $\mu_i^{(t)} = \frac{1}{|m_i^{(t)}|}\sum_{p \in m_i^{(t)}} \mathcal{Z}^{(t)}(p)$ and standard deviation. 
After \textcolor{teal}{\texttt{</dt\_rep>}}, LLM continues to generate the subsequent sequence.

\paragraph{Reasoning on Digital Twin Representation}
Following DT representation construction, LLM performs iterative reasoning using $\mathcal{D}$ as the perception of $\mathcal{X}$.
Each reasoning iteration $\mathcal{R}_i$ within tokens \textcolor{magenta}{\texttt{<think>}} and \textcolor{magenta}{\texttt{</think>}} examines specific components of $\mathcal{D}$ to extract relevant information to resolve the reasoning visual task described in the implicit text query, performing semantic reasoning related to object functions, analyzing attribute relationships, and interpreting the context of the scene.
When spatial reasoning (determining geometric relationships, relative positions, and distance calculations between objects) or temporal reasoning (analyzing motion patterns, tracking object state changes, and understanding sequential events across frames) is required, the LLM generates executable Python code $\mathcal{C}_i$ within \textcolor{violet}{\texttt{<execute>}} and \textcolor{violet}{\texttt{</execute>}} tokens.
After detection of \textcolor{violet}{\texttt{</execute>}}, LLM generates pauses and the generated code code is then operated directly on the DT representation, performing operations such as distance calculations using depth statistics $d_i^{(t)}$ and mean depth values $\mu_i^{(t)}$, geometric analysis on segmentation masks $m_i^{(t)}$, or temporal correlation analysis across frame sequences.
The results of the execution $\mathcal{O}_i$ are appended between \textcolor{violet}{\texttt{<results>}} and \textcolor{violet}{\texttt{</results>}} for the subsequent reasoning.
However, if execution errors are encountered, only relevant error information is provided (\textit{i}.\textit{e}., final line of error messages), where verbose tracebacks containing
file path information are dropped~\cite{torl}, will be appended to the \textcolor{violet}{\texttt{<results>}} and \textcolor{violet}{\texttt{</results>}}.

As complex reasoning visual tasks often require multi-step reasoning where the initial findings inform subsequent computational needs and computational results may reveal additional relationships requiring further reasoning, this reasoning process continues iteratively until the LLM determines the appropriate task type $\mathcal{T}$ within \textcolor{goldenrod}{\texttt{<task>}} and \textcolor{goldenrod}{\texttt{</task>}} tokens.
The task type $\mathcal{T}$ specifies both the category of reasoning visual task (reasoning segmentation, reasoning grounding, reasoning summarization, or reasoning visual question answering) and the corresponding output format requirements.
The final response $\mathcal{S}$ is then generated within \textcolor{goldenrod}{\texttt{<answer>}} and \textcolor{goldenrod}{\texttt{</answer>}} tokens, containing the task-specific output determined by $\mathcal{T}$. 
For reasoning segmentation, $\mathcal{S}$ includes the identified instance names and their corresponding segmentation mask file paths, while for reasoning grounding, $\mathcal{S}$ provides bounding box coordinates with instance names, for reasoning summarization or visual question answering, $\mathcal{S}$ contains natural language descriptions of relevant visual elements.

\begin{table*}[!t]
\caption{Performance comparison of video reasoning segmentation on JiTBench~\cite{jit} using region similarity ($\mathcal{J}$) and contour accuracy ($\mathcal{F}$) for semantic, spatial, and temporal reasoning at three difficulty levels.}
\label{table:jitbench}
\centering
\resizebox{\linewidth}{!}{
\begin{tabular}{l|cc|cc|cc|cc|cc|cc}
\toprule
\multirow{3}{*}{Methods} & \multicolumn{4}{c|}{Level 1 (Basic)} & \multicolumn{4}{c|}{Level 2 (Intermediate)} & \multicolumn{4}{c}{Level 3 (Advanced)} \\
\cline{2-5} \cline{6-9} \cline{10-13}
& \multicolumn{2}{c|}{Semantic / Spatial / Temporal} & \multicolumn{2}{c|}{Avg.} & \multicolumn{2}{c|}{Semantic / Spatial / Temporal} & \multicolumn{2}{c|}{Avg.} & \multicolumn{2}{c|}{Semantic / Spatial / Temporal} & \multicolumn{2}{c}{Avg.} \\
& $\mathcal{J}$ & $\mathcal{F}$ & $\mathcal{J}$ & $\mathcal{F}$ & $\mathcal{J}$ & $\mathcal{F}$ & $\mathcal{J}$ & $\mathcal{F}$ & $\mathcal{J}$ & $\mathcal{F}$ & $\mathcal{J}$ & $\mathcal{F}$ \\
\hline
LISA-7B~\cite{lisa} & 0.635 / 0.226 / 0.398 & 0.706 / 0.283 / 0.451 & 0.420 & 0.480 & 0.442 / 0.213 / 0.198 & 0.490 / 0.268 / 0.273 & 0.284 & 0.344 & 0.274 / 0.229 / 0.229 & 0.322 / 0.282 / 0.307 & 0.244 & 0.304 \\
LISA-13B~\cite{lisa} & 0.669 / 0.258 / 0.237 & 0.756 / 0.313 / 0.320 & 0.388 & 0.463 & 0.472 / 0.230 / 0.176 & 0.524 / 0.283 / 0.256 & 0.293 & 0.354 & 0.301 / 0.234 / 0.177 & 0.353 / 0.280 / 0.259 & 0.237 & 0.297 \\
GSVA~\cite{gsva} & 0.587 / 0.431 / 0.218 & 0.541 / 0.324 / 0.214 & 0.412 & 0.360 & 0.534 / 0.353 / 0.202 & 0.487 / 0.237 / 0.115 & 0.363 & 0.280 & 0.502 / 0.289 / 0.134 & 0.480 / 0.215 / 0.108 & 0.308 & 0.268 \\
LLM-Seg~\cite{llmseg} & 0.423 / 0.315 / 0.184 & 0.535 / 0.345 / 0.278 & 0.307 & 0.386 & 0.210 / 0.201 / 0.120 & 0.437 / 0.258 / 0.247 & 0.177 & 0.314 & 0.187 / 0.154 / 0.119 & 0.319 / 0.218 / 0.218 & 0.153 & 0.252 \\
V*~\cite{vstar} & 0.141 / 0.071 / 0.104 & 0.123 / 0.055 / 0.082 & 0.105 & 0.087 & 0.170 / 0.090 / 0.060 & 0.153 / 0.084 / 0.044 & 0.107 & 0.094 & 0.118 / 0.095 / 0.033 & 0.109 / 0.072 / 0.026 & 0.082 & 0.069 \\
Seg-Zero~\cite{segzero} & 0.598 / 0.445 / 0.361 & 0.622 / 0.478 / 0.334 & 0.468 & 0.453 & 0.524 / 0.398 / 0.241 & 0.548 / 0.432 / 0.309 & 0.388 & 0.430 & 0.471 / 0.352 / 0.223 & 0.523 / 0.421 / 0.281 & 0.349 & 0.408 \\
\hline
VISA-7B~\cite{visa} & 0.563 / 0.521 / 0.354 & 0.585 / 0.563 / 0.327 & 0.479 & 0.492 & 0.487 / 0.473 / 0.235 & 0.514 / 0.510 / 0.303 & 0.398 & 0.442 & 0.432 / 0.411 / 0.218 & 0.497 / 0.499 / 0.277 & 0.354 & 0.424 \\
VISA-13B~\cite{visa} & 0.599 / 0.532 / 0.389 & 0.621 / 0.587 / 0.381 & 0.503 & 0.526 & 0.521 / 0.505 / 0.267 & 0.547 / 0.543 / 0.334 & 0.431 & 0.475 & 0.463 / 0.441 / 0.248 & 0.528 / 0.531 / 0.306 & 0.384 & 0.455 \\
JiT~\cite{jit} & 0.865 / 0.789 / 0.721 & 0.795 / 0.831 / 0.793 & 0.792 & 0.806 & 0.841 / 0.752 / 0.705 & 0.801 / 0.819 / 0.784 & 0.766 & 0.801 & 0.810 / 0.741 / 0.690 & 0.801 / 0.792 / 0.737 & 0.747 & 0.777 \\
\hline
\rowcolor{gray!8}
\textbf{DT-R1} & \textbf{0.917 / 0.836 / 0.764} & \textbf{0.843 / 0.883 / 0.848} & \textbf{0.837} & \textbf{0.850} & \textbf{0.887 / 0.793 / 0.742} & \textbf{0.845 / 0.864 / 0.827} & \textbf{0.808} & \textbf{0.845} & \textbf{0.851 / 0.778 / 0.725} & \textbf{0.841 / 0.832 / 0.774} & \textbf{0.785} & \textbf{0.816} \\
\bottomrule
\end{tabular}
}
\end{table*}

\begin{table*}[!t]
\caption{Performance comparison of video reasoning segmentation methods on RVTBench~\cite{rvtbench} segmentation part.}
\label{table:rvtbench_seg}
\centering
\resizebox{\linewidth}{!}{
\begin{tabular}{l|cc|cc|cc|cc|cc|cc|cc|cc}
\toprule
\multirow{3}{*}{Methods} & \multicolumn{4}{c|}{Level 1 (Basic)} & \multicolumn{4}{c|}{Level 2 (Intermediate)} & \multicolumn{4}{c|}{Level 3 (Advanced)} & \multicolumn{4}{c}{Level 4 (Expert)} \\
\cline{2-5} \cline{6-9} \cline{10-13} \cline{14-17}
& \multicolumn{2}{c|}{Semantic / Spatial / Temporal} & \multicolumn{2}{c|}{Avg.} & \multicolumn{2}{c|}{Semantic / Spatial / Temporal} & \multicolumn{2}{c|}{Avg.} & \multicolumn{2}{c|}{Semantic / Spatial / Temporal} & \multicolumn{2}{c|}{Avg.} & \multicolumn{2}{c|}{Semantic / Spatial / Temporal} & \multicolumn{2}{c}{Avg.} \\
& $\mathcal{J}$ & $\mathcal{F}$ & $\mathcal{J}$ & $\mathcal{F}$ & $\mathcal{J}$ & $\mathcal{F}$ & $\mathcal{J}$ & $\mathcal{F}$ & $\mathcal{J}$ & $\mathcal{F}$ & $\mathcal{J}$ & $\mathcal{F}$ & $\mathcal{J}$ & $\mathcal{F}$ & $\mathcal{J}$ & $\mathcal{F}$ \\
\hline
LISA-7B~\cite{lisa} & 0.458 / 0.432 / 0.416 & 0.422 / 0.407 / 0.394 & 0.435 & 0.408 & 0.449 / 0.421 / 0.390 & 0.412 / 0.396 / 0.357 & 0.420 & 0.388 & 0.421 / 0.406 / 0.366 & 0.394 / 0.381 / 0.332 & 0.398 & 0.369 & 0.404 / 0.383 / 0.343 & 0.366 / 0.355 / 0.310 & 0.377 & 0.343 \\
LISA-13B~\cite{lisa} & 0.485 / 0.465 / 0.448 & 0.459 / 0.447 / 0.425 & 0.466 & 0.444 & 0.478 / 0.457 / 0.421 & 0.452 / 0.433 / 0.394 & 0.452 & 0.426 & 0.458 / 0.444 / 0.391 & 0.431 / 0.418 / 0.367 & 0.431 & 0.405 & 0.439 / 0.426 / 0.368 & 0.413 / 0.401 / 0.346 & 0.411 & 0.387 \\
GSVA~\cite{gsva} & 0.098 / 0.095 / 0.209 & 0.069 / 0.065 / 0.158 & 0.134 & 0.097 & 0.100 / 0.093 / 0.133 & 0.070 / 0.064 / 0.094 & 0.108 & 0.076 & 0.101 / 0.098 / 0.129 & 0.072 / 0.069 / 0.090 & 0.109 & 0.077 & 0.095 / 0.094 / 0.107 & 0.064 / 0.063 / 0.067 & 0.099 & 0.065 \\
LLM-Seg~\cite{llmseg} & 0.095 / 0.096 / 0.200 & 0.130 / 0.129 / 0.244 & 0.130 & 0.168 & 0.095 / 0.092 / 0.129 & 0.132 / 0.129 / 0.168 & 0.105 & 0.143 & 0.096 / 0.094 / 0.117 & 0.132 / 0.130 / 0.157 & 0.103 & 0.140 & 0.094 / 0.091 / 0.106 & 0.129 / 0.127 / 0.145 & 0.097 & 0.134 \\
V*~\cite{vstar} & 0.041 / 0.039 / 0.085 & 0.030 / 0.027 / 0.068 & 0.055 & 0.041 & 0.042 / 0.038 / 0.057 & 0.030 / 0.026 / 0.041 & 0.046 & 0.032 & 0.044 / 0.042 / 0.054 & 0.031 / 0.029 / 0.039 & 0.046 & 0.033 & 0.040 / 0.038 / 0.043 & 0.027 / 0.025 / 0.029 & 0.040 & 0.027 \\
Seg-Zero~\cite{segzero} & 0.479 / 0.446 / 0.413 & 0.461 / 0.437 / 0.399 & 0.446 & 0.432 & 0.459 / 0.424 / 0.387 & 0.443 / 0.416 / 0.372 & 0.423 & 0.410 & 0.438 / 0.408 / 0.365 & 0.422 / 0.398 / 0.349 & 0.404 & 0.390 & 0.413 / 0.389 / 0.347 & 0.398 / 0.375 / 0.328 & 0.383 & 0.367 \\
\hline
VISA-7B~\cite{visa} & 0.453 / 0.439 / 0.395 & 0.437 / 0.422 / 0.380 & 0.429 & 0.413 & 0.440 / 0.422 / 0.373 & 0.425 / 0.407 / 0.356 & 0.412 & 0.396 & 0.425 / 0.406 / 0.352 & 0.410 / 0.393 / 0.336 & 0.394 & 0.379 & 0.401 / 0.383 / 0.331 & 0.387 / 0.371 / 0.315 & 0.372 & 0.358 \\
VISA-13B~\cite{visa} & 0.512 / 0.488 / 0.444 & 0.491 / 0.473 / 0.426 & 0.481 & 0.463 & 0.495 / 0.472 / 0.419 & 0.477 / 0.458 / 0.401 & 0.462 & 0.445 & 0.477 / 0.456 / 0.396 & 0.461 / 0.442 / 0.378 & 0.443 & 0.427 & 0.452 / 0.431 / 0.372 & 0.436 / 0.417 / 0.354 & 0.418 & 0.402 \\
JiT~\cite{jit} & 0.548 / 0.521 / 0.475 & 0.526 / 0.506 / 0.457 & 0.515 & 0.496 & 0.529 / 0.504 / 0.448 & 0.510 / 0.490 / 0.429 & 0.494 & 0.476 & 0.510 / 0.487 / 0.423 & 0.493 / 0.472 / 0.404 & 0.474 & 0.456 & 0.483 / 0.461 / 0.399 & 0.467 / 0.447 / 0.379 & 0.448 & 0.431 \\
\hline
\rowcolor{gray!8}
\textbf{DT-R1} & \textbf{0.725 / 0.692 / 0.649} & \textbf{0.702 / 0.679 / 0.627} & \textbf{0.688} & \textbf{0.670} & \textbf{0.698 / 0.667 / 0.614} & \textbf{0.675 / 0.653 / 0.592} & \textbf{0.660} & \textbf{0.640} & \textbf{0.673 / 0.642 / 0.587} & \textbf{0.649 / 0.629 / 0.568} & \textbf{0.634} & \textbf{0.615} & \textbf{0.648 / 0.618 / 0.562} & \textbf{0.622 / 0.601 / 0.544} & \textbf{0.609} & \textbf{0.589} \\
\bottomrule
\end{tabular}
}
\end{table*}

\paragraph{Reward Design}
Following previous work \cite{r1,torl,segzero,searchr1}, our reward is based on rules to avoid the computational overhead of training neural reward models.
The total reward $R(\mathcal{Y})$ comprises one component that focuses on the formatting and another on the performance, denoted as:
\begin{equation}
R(\mathcal{Y}) = \alpha \cdot R_\text{format}(\mathcal{Y}) + \beta \cdot R_\text{accuracy}(\mathcal{Y}),
\label{eq:reward}
\end{equation}
where $\alpha$ and $\beta$ are the balancing parameters.

Format rewards $R_\text{format}(\mathcal{Y})$ evaluate the structural integrity of the rollout sequence by accessing the appearance of all required tokens in the correct order, as well as the validity of content within DT representation planning.
Formally, the format reward is computed by $R_\text{format}(\mathcal{Y}) = R_\text{token}(\mathcal{Y}) + R_\text{dag}(\mathcal{G})$, where the token format reward $R_\text{token}(\mathcal{Y})$ verifies the presence and sequential ordering of all required pair of tokens, including \textcolor{magenta}{\texttt{<think>}}, \textcolor{teal}{\texttt{<dt\_plan>}}, \textcolor{violet}{\texttt{<execute>}}, \textcolor{goldenrod}{\texttt{<task>}}, and \textcolor{goldenrod}{\texttt{<answer>}} with their corresponding closing tokens. 
The token format reward is defined as $+1$ if all tokens are correctly formatted, otherwise $-1$ to provide strong positive reinforcement for correct formatting while penalizing malformed outputs.
The DAG format reward $R_\text{dag}(\mathcal{G})$ ensures that the DT construction plan within \textcolor{teal}{\texttt{<dt\_plan>}} and \textcolor{teal}{\texttt{</dt\_plan>}} tokens forms a valid DAG by $R_\text{dag}(\mathcal{Y}) = \mathbb{I}[\text{Valid Format}] \cdot \mathbb{I}[\text{Acyclic}] \cdot \mathbb{I}[\text{Valid Dependencies}]-0.5$ where the indicators $\mathbb{I}(\cdot)$ verify that the plan is valid formatted, contains no circular dependencies, and specifies only available vision foundation models with correct dependency relationships. 
Therefore, $R_\text{dag}(\mathcal{G})$ becomes $+0.5$ if the DAG is valid otherwise $-0.5$.
We use 0.5 instead of 1.0 to provide more granular feedback, since DAG validation represents a more complex structural constraint than simple token presence, requiring the model to learn intricate relationships between vision foundation models.

Accuracy rewards $R_\text{accuracy}(\mathcal{Y})$ evaluate the performance quality of the generated output in three dimensions: (1) code executability, (2) correctness of task identification, and (3) precision of the final result, namely $R_\text{accuracy}(\mathcal{Y}) = R_\text{exec}(\mathcal{C}) + R_\text{task}(\mathcal{T}) + R_\text{result}(\mathcal{S})$. 
The code executability reward $R_\text{exec}(\mathcal{C})$ validates that Python operations within the \textcolor{violet}{\texttt{<execute>}} and \textcolor{violet}{\texttt{</execute>}} tokens can be successfully executed without syntax or run-time errors.
Following previous work \cite{torl}, we define it as $R_\text{exec}(\mathcal{C}) = 0$ if the code executes successfully, otherwise $-0.5$ to penalize non-executable operations. 
The task identification reward $R_\text{task}(\mathcal{T})$ verifies that the LLM correctly identifies the appropriate reasoning visual task type from the implicit query $Q$, providing $+0.25$ if the task specification within \textcolor{goldenrod}{\texttt{<task>}} and \textcolor{goldenrod}{\texttt{</task>}} tokens matches the ground truth task category, otherwise $0$. 
The final result accuracy reward $R_\text{result}(\mathcal{S})$ evaluates the correctness of the final output within \textcolor{goldenrod}{\texttt{<answer>}} and \textcolor{goldenrod}{\texttt{</answer>}} tokens, providing $+1$ for correct responses and $-1$ for incorrect ones to establish strong performance incentives. 
To be more concrete, for reasoning segmentation and reasoning grounding, correctness is determined by whether the IoU between the predicted and ground truth exceeds 0.5 following previous work~\cite{segzero}.
Since we do not train a segmentation decoder~\cite{lisa} with in the framework, we focus on whether the LLM successfully identifies the correct object from the DT representation by making IoU greater than 0.5 an indicator of successful object identification and localization.
For reasoning summarization and reasoning visual question answering tasks, we employ an LLM-as-a-judge~\cite{llmjudging} evaluation method that assesses the semantic correctness and factual accuracy of natural language responses generated against reference answers.

\begin{figure}[t!]
\centering
\centering
\begin{tcolorbox}[
    colback=gray!6,    
    colframe=black,      
    boxrule=0.4pt,       
    arc=0pt,             
    outer arc=0pt,       
    left=1pt,            
    right=1pt,           
    top=0pt,             
    bottom=0pt,          
    width=\linewidth 
]
\small
\textbf{DT-R1 Instruction Template}

Solve the visual reasoning task through digital twin construction and analysis. Begin by reasoning within \textcolor{magenta}{\texttt{<think>}} and \textcolor{magenta}{\texttt{</think>}} to examine the query and determine the information requirements for addressing it. Subsequently, develop a digital twin construction plan enclosed in \textcolor{teal}{\texttt{<dt\_plan>}} and \textcolor{teal}{\texttt{</dt\_plan>}} by selecting appropriate vision foundation models and defining their interdependencies using a directed acyclic graph in JSON format. You have access to SAM2 for instance segmentation, DepthAnything2 for depth estimation, Qwen2.5-VL for semantic analysis, DINO-2 for visual feature extraction, OWLv2 for object detection, and OpenCV for frame-level processing. Following plan execution, the constructed digital twin representation becomes available within \textcolor{teal}{\texttt{<dt\_rep>}} and \textcolor{teal}{\texttt{</dt\_rep>}} for subsequent analysis. Proceed with reasoning over this structured representation, utilizing iterative analysis cycles as needed. For spatial or temporal computations, implement Python code within \textcolor{violet}{\texttt{<execute>}} and \textcolor{violet}{\texttt{</execute>}}, with outcomes appearing in \textcolor{violet}{\texttt{<results>}} and \textcolor{violet}{\texttt{</results>}}. Upon completing the reasoning, specify the task category within \textcolor{goldenrod}{\texttt{<task>}} and \textcolor{goldenrod}{\texttt{</task>}}, selecting from reasoning segmentation, grounding, summarization, or visual question answering, then deliver the final answer within \textcolor{goldenrod}{\texttt{<answer>}} and \textcolor{goldenrod}{\texttt{</answer>}}.\\
\textbf{Query:} \textit{query}

\end{tcolorbox}
\caption{Prompt template for DT-R1. \textit{query} will be replaced with the actual one during training and inference.}
\label{fig:prompt}
\end{figure}

\paragraph{Training}
The training employs GRPO~\cite{grpo} to optimize LLM through the reward in Eq.~\eqref{eq:reward}.
Specifically, during each training iteration, LLM generates multiple rollout sequences by sampling the current policy $\pi_\theta$ (\textit{i}.\textit{e}., LLM), with the reward $R(\mathcal{Y})$ evaluating each sequence.
Policy updates leverage these reward signals to reinforce successful reasoning patterns while discouraging format violations and inaccurate predictions, with GRPO's variance reduction ensuring stable convergence while maintaining exploration capabilities necessary for discovering effective reasoning across diverse visual scenarios.
To prevent undesirable learning from optimizing externally generated content, including the DT representations $\mathcal{D}$ within the tokens \textcolor{teal}{\texttt{<dt\_rep>}} and \textcolor{teal}{\texttt{</dt\_rep>}} and the execution results $\mathcal{O}_i$ within the tokens \textcolor{violet}{\texttt{<results>}} and \textcolor{violet}{\texttt{</results>}}, we exclude these contexts from the training.
Finally, we utilize a standardized prompt template for both training and inference, as illustrated in Fig.~\ref{fig:prompt}.

\section{Experiments}

\paragraph{Datasets}
The training of DT-R1 requires datasets organized as triplets $(\mathcal{X}, Q, \mathcal{T}, \mathcal{S})$, where $\mathcal{X}$ represents the input visual data (image or video), $Q$ denotes the implicit text query, $\mathcal{T}$ specifies the task type, and $\mathcal{S}$ contains the ground truth output. 
The task type $\mathcal{T}$ encompasses four RVT categories: reasoning segmentation, reasoning grounding, reasoning summarization, and reasoning visual question answering, which can be derived from the dataset source itself. 
For training, we utilize four, including the training sets of ReasonSeg~\cite{lisa} for image reasoning segmentation, ReVOS~\cite{visa} for video reasoning segmentation, GroundMore~\cite{deng2025motion} for video grounding, along with 1,000 samples from ActivityNet-QA~\cite{yu2019activitynet} for video VQA and SumMe~\cite{gygli2014creating} for video summarization.
For evaluation, we utilize six datasets, categorized into in-domain and out-of-domain assessments.
In-domain evaluation utilizes test sets from training dataset sources: ReasonSeg~\cite{lisa} for image reasoning segmentation, and ReVOS~\cite{visa} test set for video reasoning segmentation. 
Out-of-domain evaluation demonstrates generalization capabilities across four datasets: LLM-Seg40K~\cite{llmseg} test set for image reasoning segmentation; JiTBench~\cite{jit,survey} for video reasoning segmentation; and RVTBench~\cite{rvtbench}, which provides evaluation across all four RVT types within the video modality.

\paragraph{Implementation Details}
We use DeepSeek-R1-Distill-Qwen-7B \cite{r1} as the backbone LLM for DT-R1 RL training.
For DT representation construction, we follow previous works~\cite{jit,rvtbench} by including SAM2~\cite{sam2}, OWLv2~\cite{owlv2}, DepthAnything~\cite{depthanything}, DINOv2~\cite{dinov2}, Qwen2.5-VL~\cite{qwen} and OpenCV operators.
All executable Python code execution utilizes Sandbox Fusion as the interpreter~\cite{torl}.
The GRPO~\cite{grpo} training configuration employs Low-Rank Adaptation (LoRA)~\cite{lora} with rank 16 to enable efficient parameter updates, with a batch size of 32 and sample number of 8.
We implement the training on 16 NVIDIA 4090 GPUs with 24GB memory using DeepSpeed~\cite{deepspeed} and trl libraries.

\begin{table*}[!t]
\caption{Performance comparison of video reasoning grounding on RVTBench-grounding~\cite{rvtbench} via cIoU and gIoU.}
\label{table:rvtbench_grounding}
\centering
\resizebox{\linewidth}{!}{
\begin{tabular}{l|cc|cc|cc|cc|cc|cc|cc|cc}
\toprule
\multirow{3}{*}{Methods} & \multicolumn{4}{c|}{Level 1 (Basic)} & \multicolumn{4}{c|}{Level 2 (Intermediate)} & \multicolumn{4}{c|}{Level 3 (Advanced)} & \multicolumn{4}{c}{Level 4 (Expert)} \\
\cline{2-5} \cline{6-9} \cline{10-13} \cline{14-17}
& \multicolumn{2}{c|}{Semantic / Spatial / Temporal} & \multicolumn{2}{c|}{Avg.} & \multicolumn{2}{c|}{Semantic / Spatial / Temporal} & \multicolumn{2}{c|}{Avg.} & \multicolumn{2}{c|}{Semantic / Spatial / Temporal} & \multicolumn{2}{c|}{Avg.} & \multicolumn{2}{c|}{Semantic / Spatial / Temporal} & \multicolumn{2}{c}{Avg.} \\
& cIoU & gIoU & cIoU & gIoU & cIoU & gIoU & cIoU & gIoU & cIoU & gIoU & cIoU & gIoU & cIoU & gIoU & cIoU & gIoU \\
\hline
LISA-7B~\cite{lisa} & 0.429 / 0.403 / 0.367 & 0.410 / 0.384 / 0.348 & 0.400 & 0.381 & 0.412 / 0.389 / 0.351 & 0.393 / 0.370 / 0.332 & 0.384 & 0.365 & 0.395 / 0.372 / 0.333 & 0.376 / 0.353 / 0.314 & 0.367 & 0.348 & 0.376 / 0.355 / 0.317 & 0.357 / 0.336 / 0.298 & 0.349 & 0.330 \\
LISA-13B~\cite{lisa} & 0.460 / 0.434 / 0.398 & 0.441 / 0.415 / 0.379 & 0.431 & 0.412 & 0.443 / 0.420 / 0.382 & 0.424 / 0.401 / 0.363 & 0.415 & 0.396 & 0.426 / 0.403 / 0.364 & 0.407 / 0.384 / 0.345 & 0.398 & 0.379 & 0.407 / 0.386 / 0.348 & 0.388 / 0.367 / 0.329 & 0.380 & 0.361 \\
GSVA~\cite{gsva} & 0.185 / 0.172 / 0.198 & 0.192 / 0.179 / 0.205 & 0.185 & 0.192 & 0.178 / 0.165 / 0.191 & 0.185 / 0.172 / 0.198 & 0.178 & 0.185 & 0.171 / 0.158 / 0.184 & 0.178 / 0.165 / 0.191 & 0.171 & 0.178 & 0.164 / 0.151 / 0.177 & 0.171 / 0.158 / 0.184 & 0.164 & 0.171 \\
LLM-Seg~\cite{llmseg} & 0.218 / 0.235 / 0.294 & 0.245 / 0.262 / 0.318 & 0.249 & 0.275 & 0.211 / 0.228 / 0.287 & 0.238 / 0.255 / 0.311 & 0.242 & 0.268 & 0.204 / 0.221 / 0.280 & 0.231 / 0.248 / 0.304 & 0.235 & 0.261 & 0.197 / 0.214 / 0.273 & 0.224 / 0.241 / 0.297 & 0.228 & 0.254 \\
V*~\cite{vstar} & 0.065 / 0.071 / 0.089 & 0.072 / 0.078 / 0.096 & 0.075 & 0.082 & 0.061 / 0.067 / 0.085 & 0.068 / 0.074 / 0.092 & 0.071 & 0.078 & 0.058 / 0.064 / 0.082 & 0.065 / 0.071 / 0.089 & 0.068 & 0.075 & 0.055 / 0.061 / 0.079 & 0.062 / 0.068 / 0.086 & 0.065 & 0.072 \\
Seg-Zero~\cite{segzero} & 0.384 / 0.362 / 0.328 & 0.395 / 0.373 / 0.339 & 0.358 & 0.369 & 0.371 / 0.349 / 0.315 & 0.382 / 0.360 / 0.326 & 0.345 & 0.356 & 0.358 / 0.336 / 0.302 & 0.369 / 0.347 / 0.313 & 0.332 & 0.343 & 0.345 / 0.323 / 0.289 & 0.356 / 0.334 / 0.300 & 0.319 & 0.330 \\
\hline
VISA-7B~\cite{visa} & 0.318 / 0.294 / 0.267 & 0.325 / 0.301 / 0.274 & 0.293 & 0.300 & 0.305 / 0.281 / 0.254 & 0.312 / 0.288 / 0.261 & 0.280 & 0.287 & 0.292 / 0.268 / 0.241 & 0.299 / 0.275 / 0.248 & 0.267 & 0.274 & 0.279 / 0.255 / 0.228 & 0.286 / 0.262 / 0.235 & 0.254 & 0.261 \\
VISA-13B~\cite{visa} & 0.491 / 0.465 / 0.429 & 0.472 / 0.446 / 0.410 & 0.462 & 0.443 & 0.474 / 0.451 / 0.413 & 0.456 / 0.432 / 0.394 & 0.446 & 0.427 & 0.457 / 0.434 / 0.395 & 0.438 / 0.415 / 0.376 & 0.429 & 0.410 & 0.438 / 0.417 / 0.379 & 0.419 / 0.398 / 0.360 & 0.411 & 0.392 \\
JiT~\cite{jit} & 0.518 / 0.492 / 0.456 & 0.499 / 0.473 / 0.437 & 0.489 & 0.470 & 0.501 / 0.479 / 0.440 & 0.483 / 0.459 / 0.421 & 0.473 & 0.454 & 0.484 / 0.461 / 0.422 & 0.465 / 0.442 / 0.403 & 0.456 & 0.437 & 0.467 / 0.444 / 0.405 & 0.448 / 0.425 / 0.386 & 0.439 & 0.420 \\
GroundMore~\cite{deng2025motion} & 0.542 / 0.518 / 0.485 & 0.524 / 0.500 / 0.467 & 0.515 & 0.497 & 0.526 / 0.504 / 0.469 & 0.508 / 0.486 / 0.451 & 0.500 & 0.482 & 0.509 / 0.487 / 0.452 & 0.491 / 0.469 / 0.434 & 0.483 & 0.465 & 0.492 / 0.470 / 0.435 & 0.474 / 0.452 / 0.417 & 0.466 & 0.448\\
\hline
\rowcolor{gray!8}
\textbf{DT-R1} & \textbf{0.699 / 0.673 / 0.638} & \textbf{0.680 / 0.654 / 0.619} & \textbf{0.670} & \textbf{0.651} & \textbf{0.674 / 0.651 / 0.615} & \textbf{0.656 / 0.632 / 0.597} & \textbf{0.647} & \textbf{0.628} & \textbf{0.653 / 0.628 / 0.594} & \textbf{0.634 / 0.609 / 0.575} & \textbf{0.625} & \textbf{0.606} & \textbf{0.629 / 0.606 / 0.571} & \textbf{0.610 / 0.587 / 0.552} & \textbf{0.602} & \textbf{0.583} \\
\bottomrule
\end{tabular}
}
\end{table*}

\begin{table*}[!t]
\caption{Performance comparison on reasoning VQA and summarization tasks with RVTBench~\cite{rvtbench} across BLEU-4~\cite{bleu}, ROUGE-L~\cite{rouge}, BertScore~\cite{bertscore}, and CIDEr\cite{cider}.}
\label{table:rvtbench_vqa_summary}
\centering
\resizebox{\linewidth}{!}{
\begin{tabular}{l|l|cccc|cccc|cccc|cccc}
\toprule
\multirow{2}{*}{\textbf{Task}} & \multirow{2}{*}{\textbf{Methods}} & \multicolumn{4}{c|}{\textbf{Level 1 (Basic)}} & \multicolumn{4}{c|}{\textbf{Level 2 (Intermediate)}} & \multicolumn{4}{c|}{\textbf{Level 3 (Advanced)}} & \multicolumn{4}{c}{\textbf{Level 4 (Expert)}} \\
\cline{3-6} \cline{7-10} \cline{11-14} \cline{15-18}
& & BLEU-4 & ROUGE-L & BertScore & CIDEr & BLEU-4 & ROUGE-L & BertScore & CIDEr & BLEU-4 & ROUGE-L & BertScore & CIDEr & BLEU-4 & ROUGE-L & BertScore & CIDEr \\
\midrule
\multirow{12}{*}{\rotatebox{90}{\textbf{VQA}}} 
& GPT-4o-mini & 0.225 / 0.163 / 0.090 & 0.300 / 0.294 / 0.204 & 0.743 / 0.729 / 0.692 & 0.708 / 0.469 / 0.264 & 0.115 / 0.113 / 0.045 & 0.221 / 0.219 / 0.193 & 0.716 / 0.714 / 0.693 & 0.500 / 0.433 / 0.104 & 0.080 / 0.042 / 0.000 & 0.202 / 0.190 / 0.268 & 0.690 / 0.690 / 0.722 & 0.255 / 0.097 / 0.561 & 0.121 / 0.058 / 0.039 & 0.213 / 0.196 / 0.187 & 0.713 / 0.688 / 0.693 & 0.312 / 0.204 / 0.109 \\
& Gemini-2.0-flash-lite & 0.113 / 0.154 / 0.045 & 0.213 / 0.230 / 0.149 & 0.653 / 0.640 / 0.580 & 0.341 / 0.338 / 0.063 & 0.076 / 0.063 / 0.020 & 0.172 / 0.174 / 0.144 & 0.632 / 0.627 / 0.544 & 0.247 / 0.212 / 0.013 & 0.037 / 0.020 / 0.000 & 0.146 / 0.140 / 0.222 & 0.579 / 0.542 / 0.650 & 0.056 / 0.011 / 0.470 & 0.033 / 0.030 / 0.013 & 0.160 / 0.146 / 0.135 & 0.604 / 0.573 / 0.538 & 0.150 / 0.051 / 0.009 \\
& Claude3-Haiku & 0.094 / 0.083 / 0.036 & 0.185 / 0.194 / 0.140 & 0.654 / 0.654 / 0.627 & 0.096 / 0.008 / 0.010 & 0.066 / 0.060 / 0.035 & 0.164 / 0.163 / 0.140 & 0.651 / 0.649 / 0.625 & 0.088 / 0.054 / 0.002 & 0.035 / 0.033 / 0.000 & 0.137 / 0.138 / 0.126 & 0.624 / 0.623 / 0.623 & 0.008 / 0.001 / 0.000 & 0.027 / 0.033 / 0.021 & 0.144 / 0.135 / 0.129 & 0.653 / 0.623 / 0.620 & 0.013 / 0.005 / 0.000 \\
& Qwen2.5-omni~\cite{Qwen2.5-Omni} & 0.089 / 0.128 / 0.044 & 0.177 / 0.207 / 0.129 & 0.617 / 0.624 / 0.579 & 0.061 / 0.081 / 0.012 & 0.048 / 0.059 / 0.042 & 0.134 / 0.135 / 0.139 & 0.596 / 0.595 / 0.585 & 0.034 / 0.038 / 0.004 & 0.040 / 0.041 / 0.000 & 0.127 / 0.136 / 0.105 & 0.576 / 0.584 / 0.557 & 0.013 / 0.004 / 0.000 & 0.053 / 0.045 / 0.035 & 0.130 / 0.124 / 0.131 & 0.602 / 0.576 / 0.584 & 0.019 / 0.009 / 0.002 \\
& Janus-Pro-7B~\cite{janus-pro} & 0.292 / 0.352 / 0.147 & 0.284 / 0.300 / 0.197 & 0.723 / 0.713 / 0.675 & 0.745 / 0.627 / 0.237 & 0.172 / 0.187 / 0.130 & 0.216 / 0.216 / 0.199 & 0.700 / 0.699 / 0.667 & 0.544 / 0.536 / 0.110 & 0.130 / 0.118 / 0.000 & 0.192 / 0.194 / 0.206 & 0.671 / 0.665 / 0.706 & 0.222 / 0.103 / 0.225 & 0.115 / 0.123 / 0.106 & 0.198 / 0.191 / 0.189 & 0.690 / 0.667 / 0.663 & 0.354 / 0.177 / 0.093 \\
& LISA-7B~\cite{lisa} & 0.266 / 0.232 / 0.082 & 0.250 / 0.243 / 0.145 & 0.700 / 0.681 / 0.619 & 0.743 / 0.511 / 0.130 & 0.126 / 0.152 / 0.058 & 0.192 / 0.193 / 0.146 & 0.672 / 0.672 / 0.630 & 0.466 / 0.458 / 0.007 & 0.070 / 0.054 / 0.000 & 0.142 / 0.143 / 0.151 & 0.617 / 0.629 / 0.668 & 0.116 / 0.007 / 0.197 & 0.058 / 0.077 / 0.043 & 0.159 / 0.146 / 0.140 & 0.661 / 0.626 / 0.628 & 0.289 / 0.103 / 0.007 \\
& LISA-13B~\cite{lisa} & 0.272 / 0.282 / 0.088 & 0.251 / 0.236 / 0.155 & 0.708 / 0.681 / 0.637 & 0.739 / 0.496 / 0.160 & 0.164 / 0.190 / 0.039 & 0.198 / 0.198 / 0.137 & 0.686 / 0.684 / 0.622 & 0.534 / 0.518 / 0.005 & 0.074 / 0.039 / 0.000 & 0.150 / 0.135 / 0.186 & 0.633 / 0.622 / 0.690 & 0.148 / 0.004 / 0.232 & 0.080 / 0.069 / 0.028 & 0.177 / 0.149 / 0.131 & 0.682 / 0.629 / 0.621 & 0.337 / 0.091 / 0.005 \\
& LISA++~\cite{lisa++} & 0.250 / 0.303 / 0.091 & 0.259 / 0.284 / 0.162 & 0.706 / 0.701 / 0.638 & 0.729 / 0.559 / 0.181 & 0.124 / 0.119 / 0.049 & 0.192 / 0.190 / 0.147 & 0.677 / 0.677 / 0.621 & 0.459 / 0.428 / 0.041 & 0.075 / 0.046 / 0.000 & 0.156 / 0.145 / 0.164 & 0.634 / 0.620 / 0.677 & 0.158 / 0.038 / 0.286 & 0.059 / 0.060 / 0.033 & 0.164 / 0.147 / 0.139 & 0.664 / 0.626 / 0.619 & 0.273 / 0.119 / 0.041 \\
& Video-R1-7B~\cite{video-r1} & 0.612 / 0.578 / 0.495 & 0.385 / 0.372 / 0.326 & 0.815 / 0.798 / 0.762 & 0.892 / 0.834 / 0.756 & 0.564 / 0.532 / 0.448 & 0.359 / 0.347 / 0.312 & 0.798 / 0.776 / 0.745 & 0.834 / 0.786 / 0.698 & 0.498 / 0.456 / 0.394 & 0.334 / 0.318 / 0.345 & 0.762 / 0.741 / 0.758 & 0.756 / 0.678 / 0.712 & 0.462 / 0.428 / 0.378 & 0.327 / 0.312 / 0.302 & 0.758 / 0.741 / 0.726 & 0.712 / 0.678 / 0.652 \\
& TinyLLaVA-Video-R1~\cite{tinyllava-video-r1} & 0.495 / 0.468 / 0.401 & 0.349 / 0.338 / 0.296 & 0.785 / 0.768 / 0.734 & 0.856 / 0.801 / 0.724 & 0.456 / 0.431 / 0.363 & 0.326 / 0.315 / 0.283 & 0.768 / 0.747 / 0.717 & 0.801 / 0.754 / 0.671 & 0.403 / 0.369 / 0.319 & 0.303 / 0.289 / 0.314 & 0.734 / 0.713 / 0.730 & 0.724 / 0.651 / 0.684 & 0.374 / 0.347 / 0.306 & 0.297 / 0.283 / 0.275 & 0.730 / 0.713 / 0.699 & 0.684 / 0.651 / 0.626 \\
\rowcolor{gray!8}& \textbf{DT-R1} & \textbf{0.845 / 0.792 / 0.678} & \textbf{0.482 / 0.479 / 0.422} & \textbf{0.872 / 0.852 / 0.826} & \textbf{0.985 / 0.918 / 0.847} & \textbf{0.781 / 0.734 / 0.612} & \textbf{0.449 / 0.437 / 0.403} & \textbf{0.852 / 0.832 / 0.818} & \textbf{0.918 / 0.867 / 0.796} & \textbf{0.695 / 0.623 / 0.547} & \textbf{0.418 / 0.397 / 0.432} & \textbf{0.826 / 0.799 / 0.815} & \textbf{0.847 / 0.748 / 0.789} & \textbf{0.642 / 0.585 / 0.518} & \textbf{0.409 / 0.390 / 0.378} & \textbf{0.815 / 0.799 / 0.789} & \textbf{0.789 / 0.748 / 0.722} \\
\midrule
\multirow{12}{*}{\rotatebox{90}{\textbf{Summary}}} 
& GPT-4o-mini & 0.047 / 0.060 / 0.076 & 0.207 / 0.211 / 0.213 & 0.687 / 0.690 / 0.692 & 0.080 / 0.075 / 0.061 & 0.061 / 0.076 / 0.042 & 0.207 / 0.209 / 0.211 & 0.691 / 0.692 / 0.696 & 0.066 / 0.060 / 0.044 & 0.074 / 0.043 / 0.025 & 0.214 / 0.212 / 0.184 & 0.693 / 0.696 / 0.672 & 0.065 / 0.047 / 0.019 & 0.088 / 0.071 / 0.040 & 0.205 / 0.207 / 0.206 & 0.686 / 0.684 / 0.689 & 0.054 / 0.029 / 0.032 \\
& Gemini-2.0-flash-lite & 0.062 / 0.059 / 0.079 & 0.187 / 0.184 / 0.199 & 0.612 / 0.613 / 0.623 & 0.036 / 0.029 / 0.016 & 0.069 / 0.071 / 0.042 & 0.186 / 0.185 / 0.175 & 0.615 / 0.613 / 0.602 & 0.027 / 0.019 / 0.003 & 0.079 / 0.042 / 0.032 & 0.199 / 0.175 / 0.178 & 0.625 / 0.603 / 0.603 & 0.017 / 0.003 / 0.006 & 0.070 / 0.076 / 0.040 & 0.178 / 0.193 / 0.174 & 0.609 / 0.618 / 0.597 & 0.017 / 0.007 / 0.003 \\
& Claude3-Haiku & 0.038 / 0.038 / 0.035 & 0.148 / 0.147 / 0.141 & 0.632 / 0.634 / 0.612 & 0.000 / 0.000 / 0.000 & 0.035 / 0.038 / 0.021 & 0.138 / 0.140 / 0.143 & 0.615 / 0.617 / 0.615 & 0.000 / 0.000 / 0.000 & 0.032 / 0.021 / 0.024 & 0.140 / 0.143 / 0.138 & 0.612 / 0.616 / 0.633 & 0.000 / 0.000 / 0.000 & 0.038 / 0.030 / 0.020 & 0.134 / 0.135 / 0.143 & 0.613 / 0.606 / 0.612 & 0.000 / 0.000 / 0.000 \\
& Qwen2.5-omni~\cite{Qwen2.5-Omni} & 0.050 / 0.052 / 0.108 & 0.170 / 0.169 / 0.191 & 0.608 / 0.609 / 0.618 & 0.019 / 0.014 / 0.019 & 0.086 / 0.087 / 0.098 & 0.174 / 0.175 / 0.201 & 0.613 / 0.612 / 0.633 & 0.016 / 0.016 / 0.005 & 0.107 / 0.098 / 0.043 & 0.191 / 0.202 / 0.163 & 0.618 / 0.633 / 0.608 & 0.021 / 0.005 / 0.008 & 0.086 / 0.103 / 0.098 & 0.167 / 0.186 / 0.201 & 0.608 / 0.614 / 0.635 & 0.006 / 0.018 / 0.005 \\
& Janus-Pro-7B~\cite{janus-pro} & 0.137 / 0.137 / 0.165 & 0.225 / 0.230 / 0.232 & 0.688 / 0.695 / 0.686 & 0.232 / 0.265 / 0.161 & 0.134 / 0.144 / 0.130 & 0.213 / 0.217 / 0.222 & 0.677 / 0.678 / 0.675 & 0.169 / 0.171 / 0.064 & 0.162 / 0.130 / 0.048 & 0.231 / 0.223 / 0.206 & 0.685 / 0.675 / 0.679 & 0.160 / 0.064 / 0.112 & 0.106 / 0.142 / 0.132 & 0.196 / 0.223 / 0.223 & 0.667 / 0.675 / 0.671 & 0.121 / 0.129 / 0.057 \\
& LISA-7B~\cite{lisa} & 0.033 / 0.052 / 0.047 & 0.119 / 0.146 / 0.117 & 0.581 / 0.613 / 0.572 & 0.041 / 0.072 / 0.008 & 0.035 / 0.040 / 0.028 & 0.110 / 0.118 / 0.086 & 0.570 / 0.578 / 0.537 & 0.022 / 0.023 / 0.000 & 0.045 / 0.029 / 0.041 & 0.112 / 0.086 / 0.148 & 0.569 / 0.537 / 0.622 & 0.009 / 0.000 / 0.042 & 0.049 / 0.053 / 0.033 & 0.133 / 0.127 / 0.095 & 0.594 / 0.581 / 0.544 & 0.025 / 0.011 / 0.000 \\
& LISA-13B~\cite{lisa} & 0.030 / 0.031 / 0.036 & 0.115 / 0.130 / 0.124 & 0.579 / 0.596 / 0.577 & 0.007 / 0.009 / 0.006 & 0.034 / 0.036 / 0.020 & 0.125 / 0.130 / 0.085 & 0.585 / 0.590 / 0.537 & 0.007 / 0.007 / 0.002 & 0.036 / 0.020 / 0.035 & 0.122 / 0.085 / 0.127 & 0.575 / 0.538 / 0.607 & 0.003 / 0.000 / 0.003 & 0.039 / 0.039 / 0.020 & 0.135 / 0.127 / 0.086 & 0.596 / 0.588 / 0.536 & 0.000 / 0.000 / 0.000 \\
& LISA++~\cite{lisa++} & 0.092 / 0.095 / 0.084 & 0.195 / 0.203 / 0.199 & 0.666 / 0.673 / 0.663 & 0.148 / 0.165 / 0.090 & 0.080 / 0.081 / 0.036 & 0.190 / 0.192 / 0.140 & 0.660 / 0.659 / 0.605 & 0.092 / 0.092 / 0.003 & 0.080 / 0.036 / 0.058 & 0.198 / 0.140 / 0.171 & 0.664 / 0.605 / 0.648 & 0.089 / 0.002 / 0.065 & 0.057 / 0.071 / 0.038 & 0.168 / 0.188 / 0.139 & 0.643 / 0.650 / 0.601 & 0.065 / 0.059 / 0.001 \\
& Video-R1-7B~\cite{video-r1} & 0.495 / 0.468 / 0.428 & 0.354 / 0.345 / 0.325 & 0.797 / 0.776 / 0.752 & 0.831 / 0.787 / 0.721 & 0.445 / 0.426 / 0.385 & 0.334 / 0.323 / 0.311 & 0.776 / 0.752 / 0.731 & 0.787 / 0.721 / 0.677 & 0.416 / 0.373 / 0.412 & 0.316 / 0.298 / 0.329 & 0.752 / 0.731 / 0.720 & 0.721 / 0.677 / 0.655 & 0.391 / 0.361 / 0.354 & 0.310 / 0.295 / 0.284 & 0.731 / 0.720 / 0.706 & 0.677 / 0.655 / 0.625 \\
& TinyLLaVA-Video-R1~\cite{tinyllava-video-r1} & 0.401 / 0.379 / 0.348 & 0.321 / 0.313 / 0.295 & 0.767 / 0.747 / 0.724 & 0.798 / 0.756 / 0.692 & 0.361 / 0.345 / 0.313 & 0.303 / 0.293 / 0.282 & 0.747 / 0.724 / 0.704 & 0.756 / 0.692 / 0.651 & 0.337 / 0.302 / 0.334 & 0.287 / 0.270 / 0.299 & 0.724 / 0.704 / 0.693 & 0.692 / 0.651 / 0.629 & 0.317 / 0.293 / 0.287 & 0.281 / 0.267 / 0.257 & 0.704 / 0.693 / 0.679 & 0.651 / 0.629 / 0.602 \\
\rowcolor{gray!8}
& \textbf{DT-R1} & \textbf{0.684 / 0.647 / 0.592} & \textbf{0.443 / 0.432 / 0.407} & \textbf{0.857 / 0.832 / 0.818} & \textbf{0.915 / 0.867 / 0.796} & \textbf{0.615 / 0.589 / 0.534} & \textbf{0.418 / 0.405 / 0.390} & \textbf{0.832 / 0.818 / 0.799} & \textbf{0.867 / 0.796 / 0.748} & \textbf{0.578 / 0.512 / 0.567} & \textbf{0.395 / 0.373 / 0.413} & \textbf{0.818 / 0.799 / 0.789} & \textbf{0.796 / 0.748 / 0.722} & \textbf{0.543 / 0.495 / 0.489} & \textbf{0.387 / 0.369 / 0.355} & \textbf{0.799 / 0.789 / 0.775} & \textbf{0.748 / 0.722 / 0.689} \\
\bottomrule
\end{tabular}
}
\end{table*}

\begin{table}[htbp]
\centering
\caption{Comparison of video reasoning segmentation on ReVoS~\cite{visa}, where DT-R1 also achieves the best performance.}
\label{tab:revos}
\resizebox{\linewidth}{!}{
\begin{tabular}{l|cc|cc|cc}
\hline
\multirow{2}{*}{Methods} & \multicolumn{2}{c|}{Referring} & \multicolumn{2}{c|}{Reasoning} & \multicolumn{2}{c}{Overall} \\
\cline{2-7}
 & $\mathcal{J}$ & $\mathcal{F}$ & $\mathcal{J}$ & $\mathcal{F}$ & $\mathcal{J}$ & $\mathcal{F}$ \\
\hline
LISA-7B~\cite{lisa} & 0.443 & 0.471 & 0.338 & 0.384 & 0.391 & 0.427 \\
LISA-13B~\cite{lisa} & 0.452 & 0.479 & 0.343 & 0.391 & 0.398 & 0.435 \\
GSVA~\cite{gsva} & 0.445 & 0.465 & 0.340 & 0.395 & 0.418 & 0.433 \\
LLM-Seg~\cite{llmseg} & 0.402 & 0.410 & 0.305 & 0.331 & 0.354 & 0.381 \\
V*~\cite{vstar} & 0.219 & 0.209 & 0.287 & 0.256 & 0.234 & 0.265 \\
Seg-Zero~\cite{segzero} & 0.487 & 0.512 & 0.361 & 0.398 & 0.424 & 0.455 \\
\hline
VISA-7B~\cite{visa} & 0.556 & 0.591 & 0.420 & 0.467 & 0.488 & 0.529 \\
VISA-13B~\cite{visa} & 0.573 & 0.608 & 0.437 & 0.481 & 0.505 & 0.545 \\
JiT~\cite{jit} & 0.691 & 0.723 & 0.618 & 0.652 & 0.655 & 0.688 \\
\hline
\rowcolor{gray!8}
\textbf{DT-R1} & \textbf{0.911} & \textbf{0.887} & \textbf{0.814} & \textbf{0.805} & \textbf{0.875} & \textbf{0.863} \\
\hline
\end{tabular}
}
\end{table}

\begin{figure}[!ht]
\centering
\includegraphics[width=\linewidth]{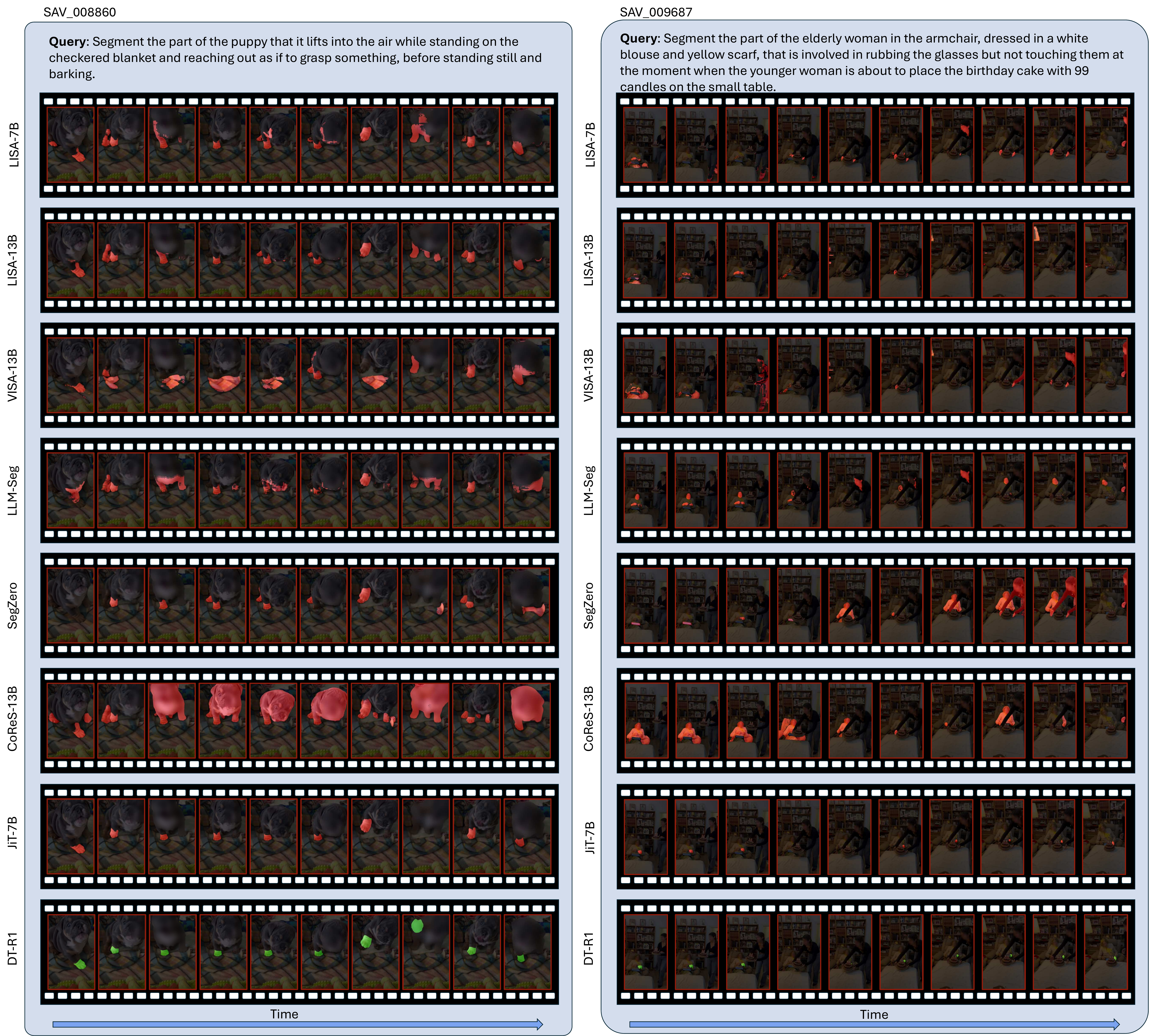}
\caption{Qualitative comparison of video reasoning segmentation results on JiTBench examples. 
Each row shows the segmentation results from different methods across time. 
Green masks indicate correct segmentation where the predicted region matches the ground truth, while red masks denote incorrect predictions. 
DT-R1 consistently produces accurate segmentations across both temporal sequences, correctly identifying the target regions throughout the video frames.
}
\label{fig:seg}
\end{figure} 

\paragraph{Evaluations of Video Reasoning Segmentation}
%
As shown in Table~\ref{table:jitbench}, DT-R1 achieves substantial improvements on JiTBench~\cite{jit}, achieving the highest $\mathcal{J}$ scores of 0.837, 0.808, and 0.785 across difficulty levels 1-3 respectively, surpassing the previous best method by 4.5\%, 4.2\%, and 3.8\%. 
Similarly, Table~\ref{table:rvtbench_seg} demonstrates DT-R1's superior performance on RVTBench-segmentation~\cite{rvtbench}, where it achieves the highest $\mathcal{J}$ scores of 0.688, 0.660, 0.634, and 0.609 across reasoning difficulty levels 1-4, with improvements of 17.3\%, 16.6\%, 16.0\%, and 16.1\% over JiT~\cite{jit}.
On ReVoS (Table~\ref{tab:revos}), DT-R1 exhibits particularly strong gains in reasoning tasks with a $\mathcal{J}$ score of 0.814 compared to JiT's 0.618, while maintaining high performance on referring tasks.
These consistent improvements across diverse benchmarks support that learning to construct and reason over DT representations through RL enables more effective preservation of spatial-temporal relationships compared to token-based approaches.

\paragraph{Evaluations of Image Reasoning Segmentation}
To evaluate DT-R1's capability in handling image-based reasoning segmentation tasks, we evaluate its performance on both in-domain and out-of-domain benchmarks. 
Table~\ref{table:image_rs} shows that DT-R1 achieves a gIoU of 0.746 and 0.812 on ReasonSeg's short and long queries respectively, representing improvements of 12.8\% and 12.9\% over the previous best method JiT. 
Importantly, DT-R1 also demonstrates generalization capabilities on the out-of-domain LLM-Seg40K~\cite{llmseg} dataset with a gIoU of 0.641, outperforming JiT's 0.485 by 15.6\%. 
These results show that DT-R1 applies to both image and video modalities.

\begin{table}[!ht]
\caption{
Performance evaluation of image reasoning segmentation on ReasonSeg~\cite{lisa} and LLM-Seg40K~\cite{llmseg} via gIoU and cIoU.
}
\label{table:image_rs}
\centering
\resizebox{\linewidth}{!}{
\begin{tabular}{l|cc|cc|cc} 
\toprule
\multirow{3}{*}{Methods} & \multicolumn{4}{c|}{ReasonSeg~\cite{lisa}} & \multicolumn{2}{c}{LLM-Seg40K~\cite{llmseg}} \\
\cline{2-5} \cline{6-7}
& \multicolumn{2}{c|}{Short Query} & \multicolumn{2}{c|}{Long Query} & \multicolumn{2}{c}{Overall} \\
\cline{2-3} \cline{4-5} \cline{6-7}
& gIoU & cIoU & gIoU & cIoU & gIoU & cIoU \\
\hline
LISA-7B~\cite{lisa} & 0.483 & 0.463 & 0.579 & 0.597 & 0.376 & 0.485 \\
LISA-13B~\cite{lisa} & 0.554 & 0.506 & 0.632 & 0.653 & 0.392 & 0.502 \\
GSVA~\cite{gsva} & 0.238 & 0.218 & 0.314 & 0.305 & 0.185 & 0.202 \\
LLM-Seg~\cite{llmseg} & 0.210 & 0.203 & 0.253 & 0.248 & 0.455 & 0.542 \\
V*~\cite{vstar} & 0.438 & 0.430 & 0.483 & 0.495 & 0.312 & 0.338 \\
Seg-Zero~\cite{segzero} & 0.531 & 0.498 & 0.587 & 0.612 & 0.442 & 0.478 \\
\hline
VISA-7B~\cite{visa} & 0.453 & 0.482 & 0.453 & 0.455 & 0.358 & 0.394 \\
VISA-13B~\cite{visa} & 0.497 & 0.521 & 0.498 & 0.502 & 0.381 & 0.417 \\
JiT~\cite{jit} & 0.618 & 0.584 & 0.683 & 0.701 & 0.485 & 0.528 \\
\hline
\rowcolor{gray!8}
\textbf{DT-R1} & \textbf{0.746} & \textbf{0.713} & \textbf{0.812} & \textbf{0.834} & \textbf{0.641} & \textbf{0.677} \\
\bottomrule
\end{tabular}
}
\end{table}

\paragraph{Evaluations of Video Reasoning Grounding}

Video reasoning (spatial-temporal) grounding aims to localize objects through bounding boxes with respect to a specific time segment of video based on given implicit text queries.
Table~\ref{table:rvtbench_grounding} reveals that DT-R1 consistently achieves the best performance across all four difficulty levels of RVTBench-grounding~\cite{rvtbench}, with cIoU scores of 0.670, 0.647, 0.625, and 0.602 from levels 1 through 4. 
These results substantially exceed the current state-of-the-art method such as GroundMore~\cite{deng2025motion}, which is specifically designed for grounding tasks.
Notably, DT-R1 maintains balanced performance across all three reasoning dimensions, achieving semantic/spatial/temporal cIoU scores of 0.699/0.673/0.638 at level 1, compared to GroundMore's 0.542/0.518/0.485.
The substantial margins of DT-R1 over both general-purpose and task-specific baselines demonstrate that our unified RL framework successfully develops grounding capabilities without requiring specialized architectural modifications or task-specific tuning.

\paragraph{Evaluations of Video Reasoning VQA and Summarization}
We evaluate DT-R1's performance on reasoning VQA and summarization on the RVTBench~\cite{rvtbench} in Table~\ref{table:rvtbench_vqa_summary}, where DT-R1 demonstrates better performance across all evaluation metrics. 
This performance advantage remains consistent as task complexity increases, with DT-R1 maintaining strong results even at the difficulty level 4, where it achieves 0.642 on BLEU-4 compared to the next best method, Video-R1-7B~\cite{video-r1}, which scores 0.462.
Recent RL-based video reasoning models, such as Video-R1-7B~\cite{video-r1}, TinyLLaVA-Video-R1~\cite{tinyllava-video-r1}, and DT-R1 demonstrate improvements over traditional approaches such as LISA~\cite{lisa} and its variants. 
For the summarization task, while overall scores are lower than VQA across all methods, DT-R1 maintains its commanding lead with scores ranging from 0.684 at the difficulty level 1 to 0.543 at level 4 for BLEU-4.
The consistent performance gap between DT-R1 and other methods across all difficulty levels underscores the effectiveness of our DT representation based reasoning framework across diverse RVTs.

\begin{table}[htbp]
\caption{Ablation study on reward components using JiTBench~\cite{jit} Level 2.}
\label{tab:ablation_reward}
\centering
\resizebox{\linewidth}{!}{
\begin{tabular}{l|cc|cc|cc|c}
\hline
\multirow{2}{*}{Configuration} & \multicolumn{2}{c|}{Format Rewards} & \multicolumn{2}{c|}{Accuracy Rewards} & \multicolumn{2}{c|}{Performance} & \multirow{2}{*}{Training Success Rate} \\
& $R_{\text{token}}$ & $R_{\text{dag}}$ & $R_{\text{exec}}$ & $R_{\text{task}}$ + $R_{\text{result}}$ & $\mathcal{J}$ & $\mathcal{F}$ & \\
\hline
w/o Format rewards & & & \checkmark & \checkmark & 0.412 & 0.438 & 31.5\% \\
w/o DAG validation & \checkmark & & \checkmark & \checkmark & 0.687 & 0.712 & 78.3\% \\
w/o Code execution & \checkmark & \checkmark & & \checkmark & 0.724 & 0.756 & 85.7\% \\
w/o Accuracy rewards & \checkmark & \checkmark & & & 0.523 & 0.548 & 62.1\% \\
Only task accuracy & & & & \checkmark & 0.298 & 0.315 & 22.8\% \\
\hline
Full DT-R1 & \checkmark & \checkmark & \checkmark & \checkmark & \textbf{0.808} & \textbf{0.845} & 94.2\% \\
\hline
\end{tabular}
}
\end{table}

\begin{figure}[htbp]
\centering
\includegraphics[width=\linewidth]{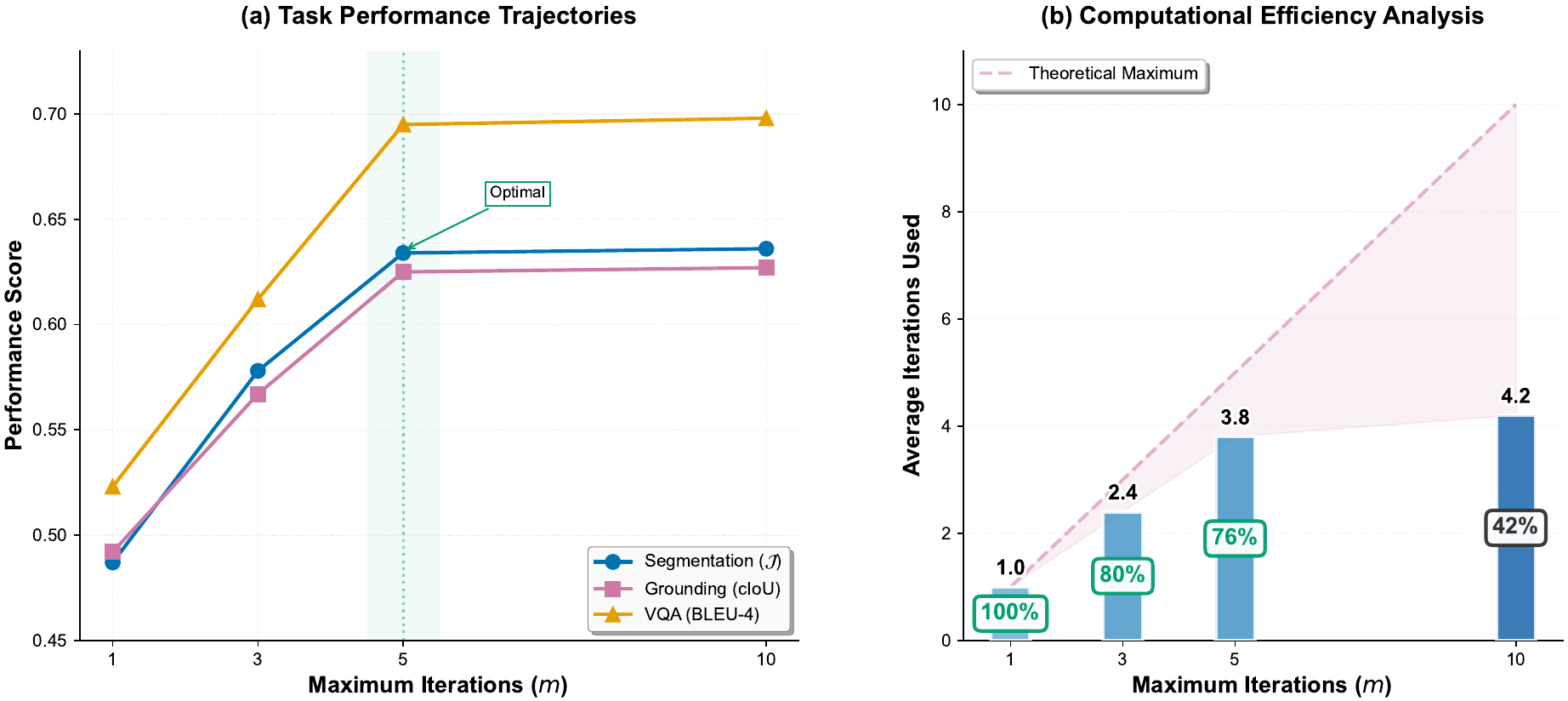} 
\caption{
Analysis of iterative reasoning cycles impact on DT-R1 performance. 
(a) It demonstrates the performance across segmentation, grounding, and visual question answering tasks as maximum iterations increase from 1 to 10, revealing performance saturation at 5 iterations. 
(b) It illustrates computational efficiency, where the average number of reasoning iterations actually utilized versus the theoretical maximum, with efficiency percentages indicating the model's adaptive termination behavior. 
}
\label{fig:ablation}
\end{figure}

\paragraph{Ablation Study}
We conduct an ablation study on the contribution of each reward component on JiTBench Level 2, as presented in Tab.~\ref{tab:ablation_reward}. 
The results demonstrate that each component of our reward function helps achieve optimal performance and training stability. 
Removing format rewards entirely leads to substantial performance degradation, with $\mathcal{J}$ and $\mathcal{F}$ scores dropping to 0.412 and 0.438 respectively, alongside a reduction in training success rate to 31.5\%. 
The DAG validation component proves particularly important, as its removal reduces performance to 0.687 $\mathcal{J}$ and 0.712 $\mathcal{F}$, while the code execution reward also contributes meaningfully to the final results. 
The configuration using only task accuracy performs poorly across all metrics (0.298 $\mathcal{J}$, 0.315 $\mathcal{F}$), highlighting the necessity of our reward design.
%
%
%
Additionally, we explore the impact of iterative reasoning cycles on performance, as shown in Fig.~\ref{fig:ablation}. 
The DR-R1 demonstrates computational efficiency by adaptively utilizing an average of 3.8 iterations when permitted up to 5, indicating that it learns to terminate reasoning cycles appropriately without exhausting the maximum allowed iterations. 

\section{Conclusion}

We present DT-R1, a RL framework that trains LLMs to construct and reason over digital twin representations as a unified solution for diverse reasoning visual tasks. 
%
%
While the performance of the DT-R1 approach seems promising, the current design requires executing multiple vision foundation models during inference for DT representation construction; hence one future direction can look into efficient implementation such as distilled models~\cite{fastsam,fastsam3d} or caching~\cite{liu2024efficient} to reduce computations. 
Additionally, DT-R1 currently processes only visual and textual modalities, yielding opportunities to incorporate audio and other sensory data into the DT representation as another future direction. 
Future work can also explore scaling to larger LLMs to enhance reasoning capabilities and enable more complex DT construction that capture complex spatial-temporal relationships.
By demonstrating that pure RL can train LLMs to show visual reasoning capabilities, DT-R1 shows a promise that can complement the current paradigm~\cite{position} of maintaining separate models for each reasoning visual task.

{
    \small
    \bibliographystyle{ieeenat_fullname}
    \bibliography{main}
}


\end{document}